\documentclass{article}

\PassOptionsToPackage{numbers}{natbib}
\usepackage[preprint]{neurips_2025}

\usepackage[utf8]{inputenc} % allow utf-8 input
\usepackage[T1]{fontenc}    % use 8-bit T1 fonts
\usepackage{hyperref}       % hyperlinks
\usepackage{url}            % simple URL typesetting
\usepackage{booktabs}       % professional-quality tables
\usepackage{amsfonts}       % blackboard math symbols
\usepackage{nicefrac}       % compact symbols for 1/2, etc.
\usepackage{microtype}      % microtypography
\usepackage{xcolor}         % colors

\usepackage{graphicx}

\title{Structure Disruption: Subverting Malicious Diffusion-Based Inpainting via Self-Attention Query Perturbation}

\author{%
	Yuhao He \\
	Faculty of Innovation Engineering\\
	Macau University of Science and Technology\\
	Macao, China \\
	\texttt{3250004430@student.must.edu.mo} \\
	\And
	Jinyu Tian \thanks{corresponding author.} \\
	Faculty of Innovation Engineering \\
        Macau University of Science and Technology\\
	Macao, China \\
	\texttt{jytian@must.edu.mo} \\
	\And
	Haiwei Wu \\
		School of Computer Science and Engineering \\
        University of Electronic Science and Technology of China\\
	Chengdu, Sichuan, China \\
        \texttt{haiweiwu@uestc.edu.cn} \\
	\And
	Jianqing Li \\
	Faculty of Innovation Engineering\\
	Macau University of Science and Technology\\
	Macao, China \\
        \texttt{jqli@must.edu.mo} \\
}

\begin{document}

\maketitle

\begin{abstract}
The rapid advancement of diffusion models has enhanced their image inpainting and editing capabilities but also introduced significant societal risks. Adversaries can exploit user images from social media to generate misleading or harmful content. While adversarial perturbations can disrupt inpainting, global perturbation-based methods fail in mask-guided editing tasks due to spatial constraints. To address these challenges, we propose \textbf{Structure Disruption Attack (SDA)}, a powerful protection framework for safeguarding sensitive image regions against inpainting-based editing. Building upon the contour-focused nature of self-attention mechanisms of diffusion models, SDA optimizes perturbations by disrupting queries in self-attention during the initial denoising step to destroy the contour generation process. This targeted interference directly disrupts the structural generation capability of diffusion models, effectively preventing them from producing coherent images. We validate our motivation through visualization techniques and extensive experiments on public datasets, demonstrating that SDA achieves state-of-the-art (SOTA) protection performance while maintaining strong robustness.
\end{abstract}

\section{Introduction}
\label{Introduction}
The rapid advancement of diffusion models has revolutionized image synthesis, facilitating the efficient generation of photorealistic and high-fidelity images \cite{dhariwal2021diffusion, rombach2022high, zhang2023adding, brooks2023instructpix2pix}. In conditional generation tasks, these models can be fine-tuned on limited exemplars to capture intricate stylistic attributes \cite{gal2023an, hu2022lora, mou2024t2i, zhang2023inversion} or reproduce specific objects and identities with remarkable precision \cite{ruiz2023dreambooth, kumari2023multi, ye2023ip, dong2024continually}. For inpainting applications \cite{lugmayr2022repaint, xie2023smartbrush, xu2024personalized, yang2023uni}, diffusion models leverage spatially constrained mask-guided synthesis, enabling targeted content generation within user-defined regions while maintaining coherence with surrounding context through text-guided conditioning.
	
\begin{figure}[t]
    \centering
    \includegraphics[width=0.9\textwidth]{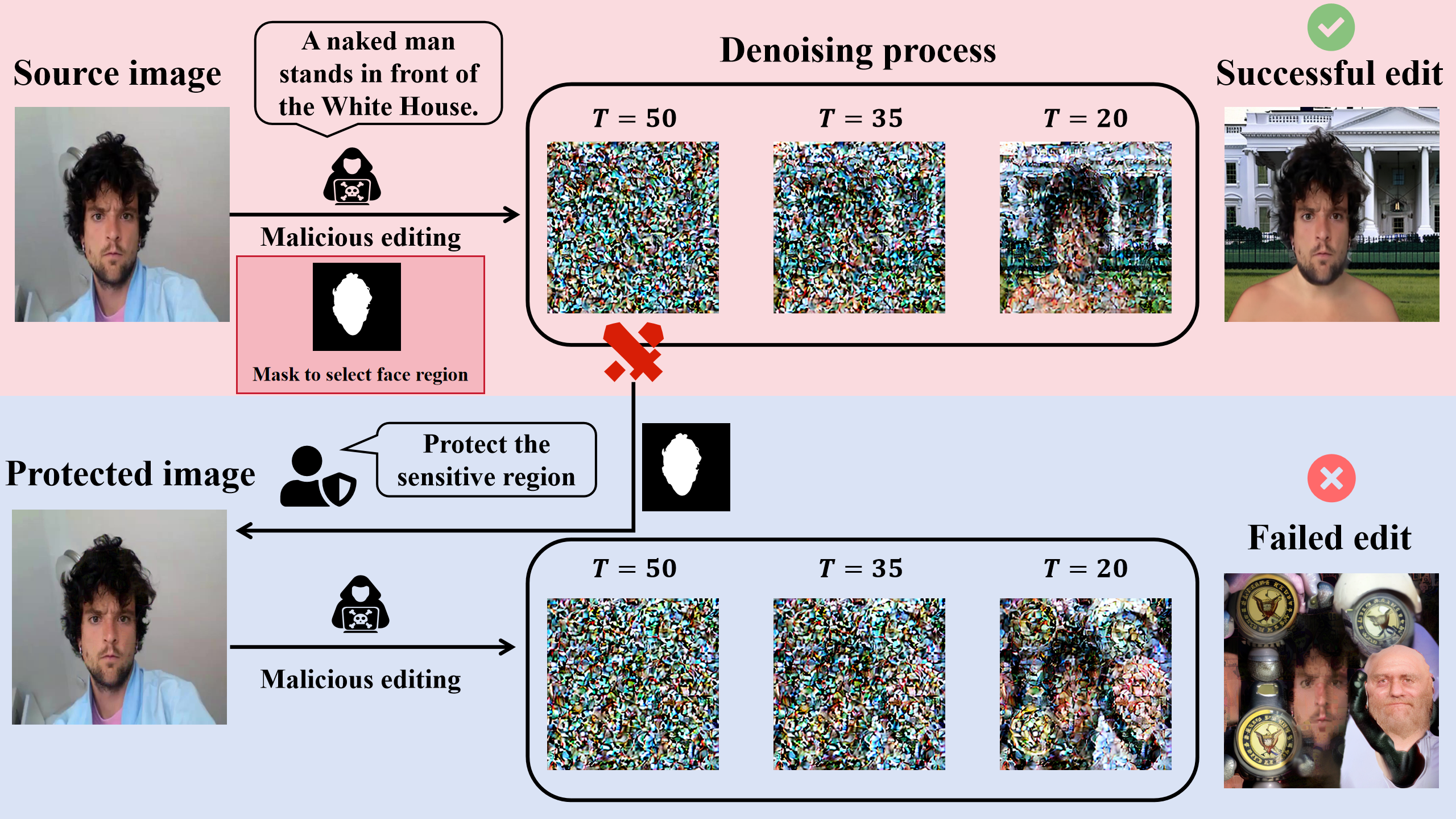}
    \caption{Protected vs. Unprotected Image Resistance: (Top) Malicious inpainting alters contextual elements (e.g. naked and the White House) while preserving key features (e.g. human face). (Bottom) SDA-protected images demonstrate robust resistance to such editings through sensitive region encryption, effectively neutralizing unauthorized edits.}
    \label{fig:overview}
\end{figure}

However, the rapid advancement of diffusion models has raised significant ethical and legal concerns \cite{lindberg2024applying, cavna2023artists}. For instance, unauthorized generation of images mimicking a specific artist’s style may constitute copyright infringement, and malicious actors can fine-tune these models to synthesize highly realistic yet fabricated portraits in diverse contexts, facilitating the creation of deepfake misinformation. Furthermore, the advancement of inpainting has lowered the technical barrier to image manipulation, potentially leading to the rampant spread of misinformation. As illustrated in the red region of Figure \ref{fig:overview}, attackers can extract user photos from social media, employ a mask to preserve facial features, and combine them with negative prompts to generate defamatory or misleading imagery, severely violating personal reputation rights.
	
To mitigate these risks, previous research has explored the use of global adversarial perturbations to disrupt the denoising process of diffusion models, demonstrating promising protection efficacy in fine-tuning-based text-to-image generation tasks \cite{van2023anti, liang2023adversarial, xue2024toward, shan2023glaze}. However, in mask-guided image editing (e.g., inpainting), only the masked regions interact with the model, rendering global perturbations ineffective. Currently, inpainting-specific protection methods remain underexplored, often exhibiting suboptimal and unstable performance \cite{Salman2023Raising, choi2025diffusionguard}. 

In this work, we present \textbf{Structure Disruption Attack (SDA)}, an innovative and efficient image protection method designed to prevent malicious editing of sensitive image region via diffusion model-based inpainting. 
% SDA optimizes adversarial perturbations by maximizing the discrepancy in self-attention queries between clean images and their adversarial counterparts during the initial step of the diffusion process, while ensuring the added perturbations remain imperceptible.
The core motivation of SDA stems from the observation that diffusion models typically generate images through a coarse-to-fine process \cite{yang2023diffusion}: \textbf{Early denoising timesteps establish object contours, while later timesteps progressively refine details. Therefore, we propose to disrupt contour structure of the protected contents (e.g. human faces), during the initial diffusion phase to prevent the synthesis of a complete image.} Since the self-attention mechanism primarily governs structural coherence \cite{liu2024towards} (e.g., texture consistency and spatial relationships) by focusing on these contours, we implement this disruption through self-attention interference during the initial denoising step. As illustrated in Figure \ref{fig:overview} (blue region), this disruption prevents the model from reconstructing primary object structures, ultimately leading to incomplete image generation. Extensive experiments in the last section would not only validate our design rationale but also demonstrate SDA's remarkable performance. In summary, our contributions are as follows:
	
\begin{itemize}
    \item We propose structural disruption as a novel defense mechanism to prevent the misuse of diffusion-based inpainting, offering targeted protection for sensitive regions.
    \item We reveal a phenomenon by visualization analysis that perturbing self-attention queries in early diffusion steps triggers a cascade failure: the model not only loses the ability to capture object contours but also breaks semantic alignment with text prompts, leading to complete generation collapse.
    \item Our method achieves state-of-the-art performance in countering diffusion-model-based inpainting misuse and demonstrates effective performance across defenses and model versions. We further validate the practical effectiveness of the proposed method in real-world scenarios simulated through mask augmentation \cite{choi2025diffusionguard}.
\end{itemize}

\section{Related work}
	
\paragraph{Adversarial examples against diffusion models} The growing adoption of diffusion models in content creation has raised critical security concerns, particularly regarding unauthorized image synthesis. Researchers have developed specialized adversarial attacks that exploit these models' noise prediction mechanisms to prevent malicious usage. AdvDM \cite{liang2023adversarial} optimizes perturbations by directly maximizing the diffusion loss, while Anti-Dreambooth \cite{van2023anti} alternately maximizes this loss and minimizes the Dreambooth \cite{ruiz2023dreambooth} training loss through joint perturbation-model optimization. CAAT \cite{xu2024perturbing} enhances this approach by fine-tuning cross-attention blocks during noise prediction maximization. Alternative protection strategies manipulate latent space encodings through encoder-targeted optimization. Mist \cite{liang2023mist} jointly optimizes noise prediction maximization and latent alignment minimization, while SDST \cite{xue2024toward} improves optimization efficiency through score distillation sampling.
	
\paragraph{Adversarial examples against inpainting} Unlike standard image generation tasks that process complete inputs, inpainting models operate under constrained conditions where only specific masked regions of the image are modifiable. This partial accessibility requirement fundamentally alters the threat model, as conventional global adversarial protection methods cannot function effectively when models inherently ignore unmasked areas. To address this challenge, researchers have developed localized attack strategies. Photoguard \cite{Salman2023Raising} proposes two localized strategies: EncoderAttack for latent space manipulation and DiffusionAttack for full-process interference. DiffusionGuard \cite{choi2025diffusionguard} introduces mask augmentation to strengthen local perturbations while attacking initial denoising steps. Advanced approaches disrupt semantic consistency through latent centroid deviation (DDD \cite{son2024disrupting}) or attention mechanism perturbation (AdvPaint \cite{jeon2025advpaint}). Our analysis reveals that targeted self-attention interference provides superior protection by directly preventing coherent image generation.
	
\section{Methodology}
	
In this section, we introduce the SDA, an efficient image protection method that prevents images from being maliciously inpainted through Stable Diffusion models. Our approach operates by \textbf{directly disrupting the contour structure during the initial phase of denoising to prevent the model from reconstructing primary object structures, this intervention fundamentally compromises the model's capacity to synthesize structurally coherent imagery.} Before discussing technical details, we present the threat model that governs our security analysis. 

% \subsection{Problem statement}
	
% Existing security approaches for stable diffusion predominantly employ \textit{global} perturbation to disrupt the denoising process directly \cite{liang2023adversarial}. However, these methods exhibit fundamental limitations in the inpainting scenario, where masked processing preserves only partial image content. This limitation arises because the localized nature of inpainting operations restricts perturbation interactions to specific subregions, effectively bypassing global protection mechanisms. Our analysis reveals this critical vulnerability necessitates developing specialized defense paradigms for partial image manipulation contexts.
	
\paragraph{Threat model} We assume that adversaries could strategically mask the sensitive image region (e.g., human face) and exploit diffusion-powered inpainting through arbitrary malicious prompts to synthesize reputationally damaging content. This threat formulation aligns with real-world adversarial patterns: sensitive regions like facial areas in portraits or primary subjects in other image types are prioritized targets due to their high privacy value and ethical impact potential \cite{pawelec2024decent}. Previous work \cite{choi2025diffusionguard} also adopts the same paradigm of the sensitive region, where targeted perturbation optimization resolves the inherent limitations of global perturbations \cite{van2023anti} in inpainting scenarios, and this constrained threat model enables us to focus our protection efforts on critical components of the image while maintaining practical applicability. 

% Existing security approaches for Stable Diffusion predominantly employ \textit{global} perturbations to disrupt denoising \cite{liang2023adversarial}, yet fail in inpainting scenarios where localized operations restrict perturbations to specific subregions, inherently bypassing global protections. SDA selectively safeguards the \textbf{sensitive} region within the image through targeted protection.

% \paragraph{Threat model} We posit that malicious actors could strategically mask sensitive image regions (e.g., face) and subsequently exploit text-guided inpainting diffusion model through adversarial prompts to synthesize deceptive or harmful visual content. Although adversaries possess the theoretical capability to manipulate arbitrary image regions, achieving universal protection against all possible mask configurations presents fundamental technical barriers. According to prior work \cite{choi2025diffusionguard}, we therefor propose a pragmatic framework based on the observation that attackers typically target semantically \textbf{sensitive} regions, facial areas in portraits or primary subjects in other image types. This constrained threat model enables us to focus protection efforts on critical image components while maintaining practical applicability. The core technical implementation follows.

\subsection{Adversarial self-attention disruption in initial diffusion step}
We begin our discussion of SDA's technical details by first analyzing why perturbing the initial denoising step effectively prevents malicious image inpainting. Following this analysis, we present our methodology for executing the perturbation process.

\begin{figure}[t]
    \centering
    \includegraphics[width=0.9\textwidth]{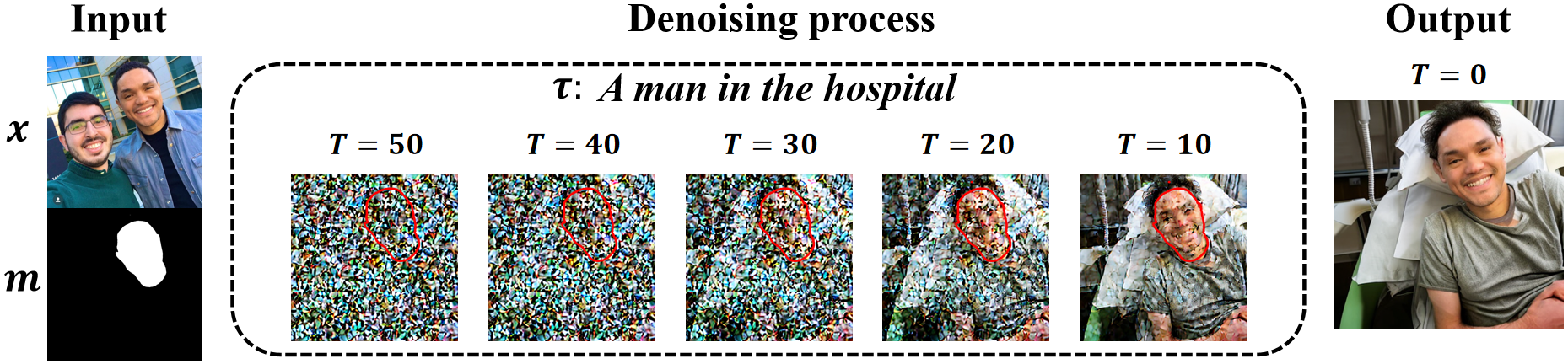}
    \caption{Denoising process of inpainting diffusion models. We visualize intermediate denoising process outputs and use red curves to mark the facial contours during the denoising process.}
    \label{fig:process}
\end{figure}

\paragraph{Why perturbing the initial denoising step effective} The denoising trajectory of diffusion models inherently follows a coarse-to-fine generation pattern \cite{NEURIPS2020_4c5bcfec,yang2023diffusion}. Empirical observations demonstrate that in the early steps ($T$ is large), the model prioritizes establishing low-frequency components that define global semantics (e.g. contour, composition). Subsequent steps ($T$ is small) gradually inject high-frequency elements to refine textures (e.g., edges and local details). Figure \ref{fig:process} corroborates this statement, as time step $T$ decreases from $50$ to $20$, the image's contours become increasingly clear. Then, when $T$ reaches 10, the details of the image are enriched. Motivated by this observation, we propose the SDA to disrupt the contour structure during the initial phase of denoising. Moreover, our SDA focus on the initial denoising step substantially reduces both computational overhead and temporal expenditure in perturbation optimization compared to full-chain adversarial attacks which typically require the backpropagation of the full generation steps. 

\paragraph{How to disrupt the contour during the initial denoising step} The self-attention mechanism are crucail in stable diffusion models. It primarily govern the structural contour formation in image synthesis, where targeted disruption can erase critical object semantics and collapse the generative integrity. The attention mechanism enables diffusion models to adaptively modulate their focus across different spatial regions during the denoising process, which typically consists of self-attention and cross-attention. Self-attention models geometric topology and structural priors (e.g., object contours, texture continuity) by aggregating spatial correlations within latent representations to ensure visual coherence in the output \cite{liu2024towards}. Cross-attention achieves fine-grained fusion of cross-modal features by computing interaction weights between the image latent space and the semantic text space, ensuring semantic consistency between generated content and textual descriptions \cite{liu2024towards, saharia2022photorealistic}.

\begin{figure}[t]
    \centering
    \includegraphics[width=0.8\textwidth]{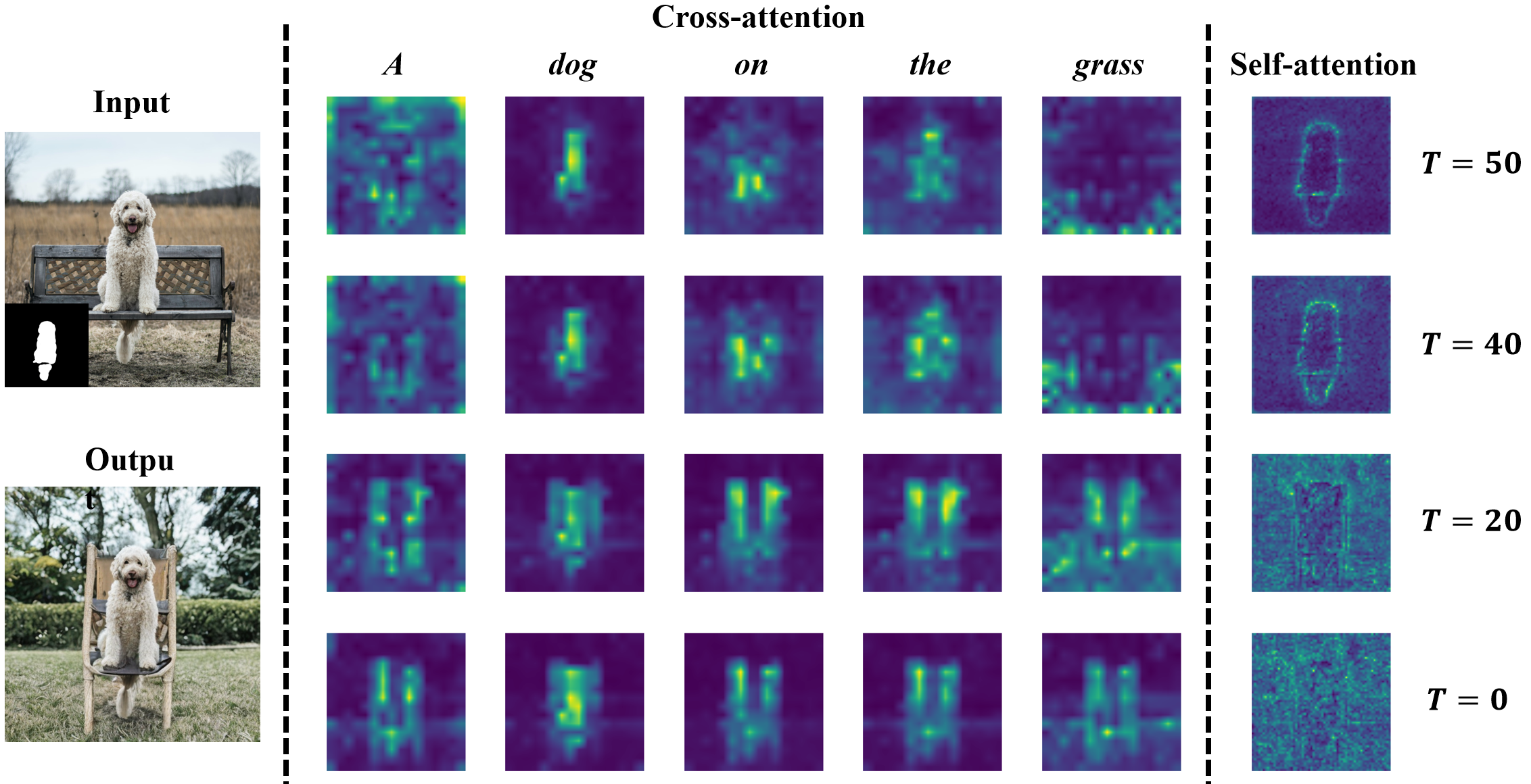}
    \caption{The attention map during the denoising process of Inpainting}
    \label{fig:attention_map}
\end{figure}

Figure \ref{fig:attention_map} illustrates the operational mechanism of attention during the denoising process. The cross-attention module primarily facilitates the semantic interaction between the image and the text prompt. For instance, when generating a "dog", the attention weights predominantly focus on canine features, while shifting to ground regions when generating "grass". In contrast, the self-attention mechanism mainly focuses on object contours and maintains global structural consistency throughout the image. Building upon these insights and integrating previous findings, our proposed SDA strategically targets the self-attention mechanism during the initial denoising phase. By disrupting the model's structural comprehension at this critical stage, SDA triggers a cascading effect that ultimately prevents the generation of coherent images.

Attention mechanism can be formally expressed as follows:
	
\begin{equation}
    \label{eq:attention}
    Attention(Q,K,V) = Softmax\Big(\frac{QK^{T}}{\sqrt{d}}\Big)\cdot V ,
\end{equation}
	
$d$ is the dimension of $Q$ and $K$. Here, $Q = q(\phi(I))$ represents the query, where $\phi(\cdot)$ indicates the latent space mapping of the input $I$ from the previous layer before the attention block, and $q(\cdot)$ is the linear projection operator for $Q$ \cite{jaegle2021perceiver, vaswani2017attention}. Similarly, $K$ and $V$ represent the key and the value, respectively. $Q$, $K$, and $V$ are the core components of the attention mechanism. In self-attention, $Q$, $K$, and $V$ are derived from the latent space of images while in cross-attention, $Q$ is derived from the image and $K$ and $V$ originate from the text. We define the adversarial objective as follows:
	
\begin{equation}
    \label{eq:sda}
    \delta = \mathop{\arg\max}_{||\delta||_{2}\leq\eta}\sum_{l} ||\hat{Q_{s}^{l}} - Q_{s}^{l}||,
\end{equation}
	
$Q_{s}^{l} = q_{s}^{l}(\phi_{T}^{l}(I))$ and $\hat{Q_{s}^{l}} = q_{s}^{l}(\phi_{T}^{l}(I+\delta))$, where $s$ represents self-attention, $T$ is the initial step of denoising, $l$ means in the $l$-th layer of U-Net and $\delta \in \mathbb{R}^{H\times W\times 3}$ denotes the protective perturbation constrained by $\Vert \delta \Vert_{2} \leq \eta$ (where $\eta > 0$) to be optimized, where $H$ and $W$ are the height and the width of the origin image respectively. Eq. \ref{eq:sda} optimizes the protective perturbation by maximizing the discrepancy of queries in self-attention during the initial step of the diffusion model's denoising process. This interference disrupts the model's holistic perception of the image, thereby triggering a chain reaction that prevents the generation of a complete image.
\begin{figure}[t]
    \centering
    \includegraphics[width=0.9\textwidth]{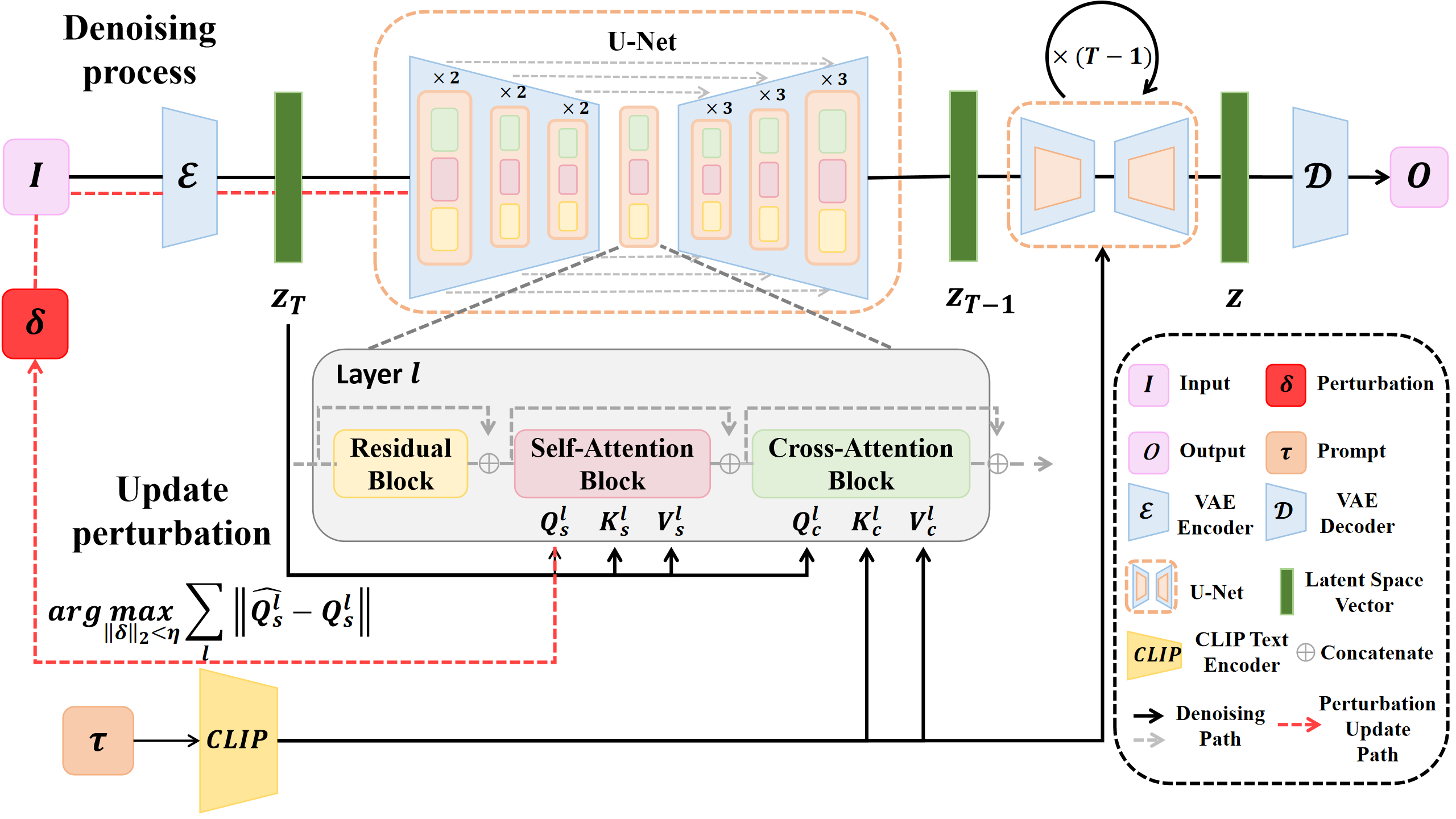}
    \caption{The inpainting diffusion pipeline and the protective perturbation update process. The black path is the inpainting denoising process, and the red path is the perturbation update process.}
    \label{fig:U-Net}
\end{figure}
The optimization objective (\ref{eq:sda}) further demonstrates the computational efficiency of our SDA framework, as it requires only performing forward and backward propagation through the initial denoising step rather than computing the full iterative chain, with detailed complexity analysis provided in the Appendix.

Figure \ref{fig:U-Net} shows the inpainting diffusion pipeline and the SDA pipeline, where $I$ is the input, and the output $O$ is the restored image. The black path represents the inpainting diffusion pipeline, which initially employs a VAE \cite{kingma2013auto} encoder to map the input into the latent space. Through $T$ iterative steps, a U-Net architecture progressively predicts and removes noise at each stage, ultimately reconstructing the final output image via the VAE decoder. As the core denoising operator in diffusion models, the U-Net architecture hierarchically integrates three principal components: (i) residual blocks with skip connections for feature preservation, (ii) multi-head self-attention mechanisms for contextual modeling, and (iii) cross-attention modules enabling conditional guidance through auxiliary inputs. The red path delineates the perturbation update process. During the initial denoising step, the SDA extracts query vectors from the self-attention block and optimizes perturbations through Eq. \ref{eq:sda}. 

\begin{figure}[t]
    \centering
    \includegraphics[width=0.8\textwidth]{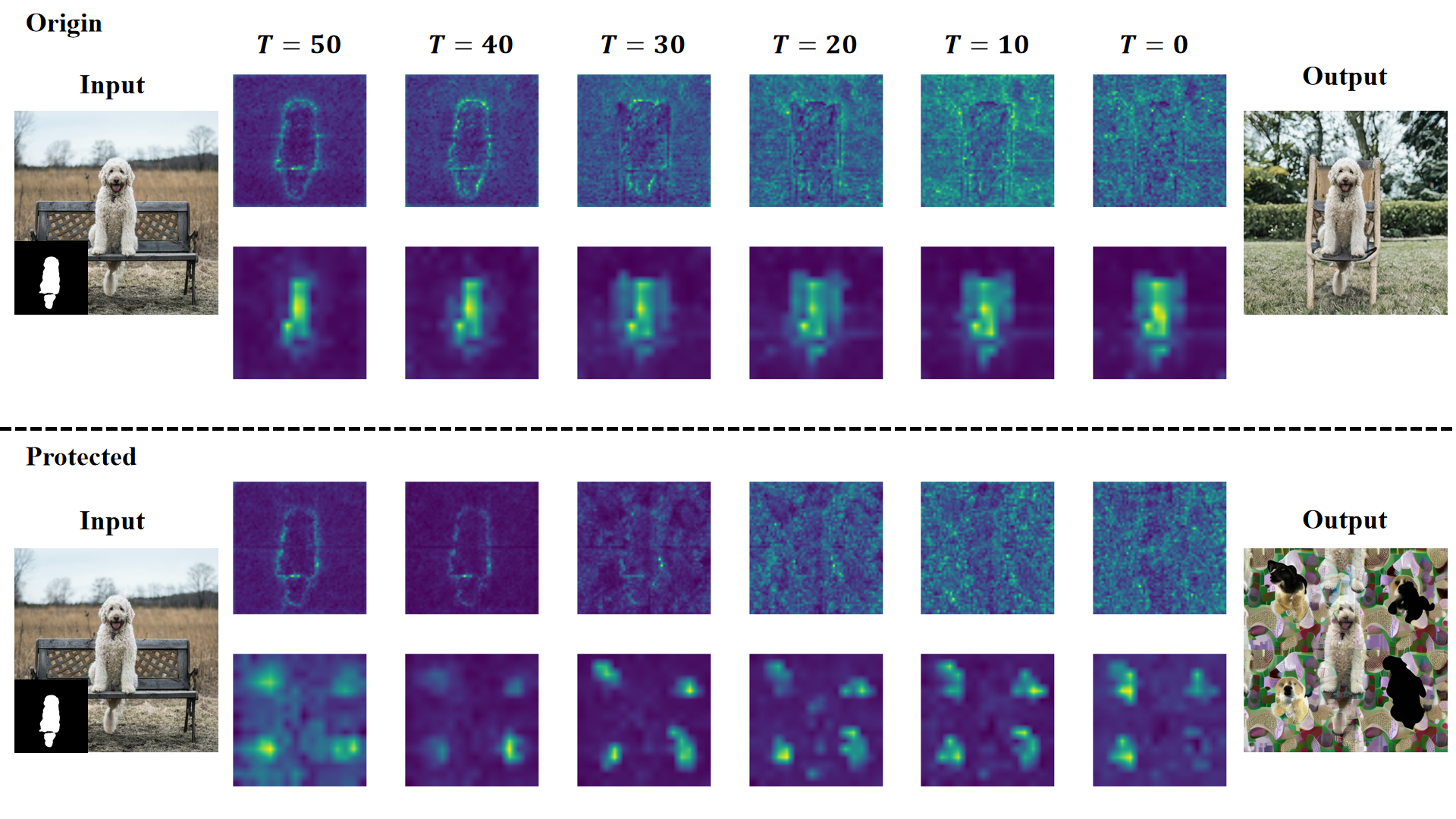}
    \caption{Comparison of attention maps during the generation process between original and protected images. Above the dashed line: generation process of the original image, where the first row shows self-attention maps and the second row displays cross-attention maps for the "dog" token. Below the dashed line: generation process of the image protected by SDA.}
    \label{fig:compare_attention}
\end{figure}
	
\subsection{Empirical analysis}
Before delving into the discussion of the related experimental results, we conduct a brief empirical analysis of SDA. Figure \ref{fig:compare_attention} compares the attention maps during the generation process of original and protected images. Observing the self-attention maps of the original image, we can identify that during the initial denoising phase (when $T$ is relatively large), the attention primarily focuses on the contours of the main subject, subsequently diffusing outward (as $T$ gradually decreases). This pattern indicates that the model initially concentrates on the overall composition of the image before enriching other details. In contrast, the self-attention maps of the protected image reveal a significant reduction in the model's attention to the image contours. When $T=40$, the model almost loses its ability to model the contours of the image, resulting in the failure to capture crucial information about the overall composition. This impairment leads to the loss of detail generation capability, as evidenced at $T=30$, triggering a chain reaction that ultimately prevents the generation of a complete image. This phenomenon aligns consistently with our initial motivation. An examination of the cross-attention maps further reveals that the chain reaction initiated by the interference with self-attention during the initial denoising phase also disrupts the alignment capability between cross-attention and the reference text. As evidenced at $T=50$, the model loses its ability to align the main subject of the image with the textual token "dog".
	
\section{Experiments}
In this section, we present a comparative evaluation of SDA against state-of-the-art protect methods for inpainting. We further investigate the transferability of SDA in black-box settings and assess its robustness. Finally, we employ mask augmentation \cite{choi2025diffusionguard} to simulate real-world deployment scenarios, systematically evaluating the robustness of protection methods under diverse masking conditions.
	
\subsection{Experimental setup}
\label{Experimental setup}
All adversarial attacks in our experiments are restricted by the budget of $\eta = 12$ and $300$ iterations for optimization. We use the Nvidia A6000 48GB GPU, which took less than 1.5 minutes to optimize per image. We evaluate our proposed method on the pre-trained inpainting model from Stable Diffusion v2 in this experiment. 
	
\paragraph{Datasets} We constructed two distinct datasets to evaluate different scenarios: (1) face dataset and (2) instance dataset. The face dataset was compiled from publicly available online sources \cite{MultiClassFaceSegmentation}, where we randomly sampled 100 image-mask pairs following previous work \cite{Salman2023Raising}. For the instance dataset, we use the COCO benchmark \cite{lin2014microsoft}: we randomly select 10 categories from COCO, with 10 image-mask pairs per category (totaling 100 samples). Then we employed ChatGPT \cite{chatgpt2025} to generate the corresponding textual prompts (e.g., "A man in the hospital", "A bear in the forest") for each image-mask pair. Crucially, we assigned a unique fixed random seed to each image-mask pair to ensure rigorous experimental control and enable exact reproducibility across comparisons. All images are resized to $512\times512$.
	
\paragraph{Metric} We employ comprehensive metrics for quantitative analysis: reference-based metrics (VIF \cite{sheikh2006image}, SSIM \cite{wang2004image}, PSNR \cite{jahne2005digital}, FID \cite{heusel2017gans}, LPIPS \cite{zhang2018unreasonable}) compare protected inpainting results against baseline inpainting results (generated from original images), and non-reference metrics (CLIP Score \cite{hessel2021clipscore} for prompt-image alignment, PIQE \cite{mittal2012no} for perceptual quality assessment). Each test case maintains strict one-to-one correspondence between image, mask, prompt and random seed to ensure evaluation fairness, as specified in our experimental design.
	
\subsection{Comparison with existing methods on inpainting tasks}
	
\begin{figure}[t]
    \centering
    \includegraphics[width=0.8\textwidth]{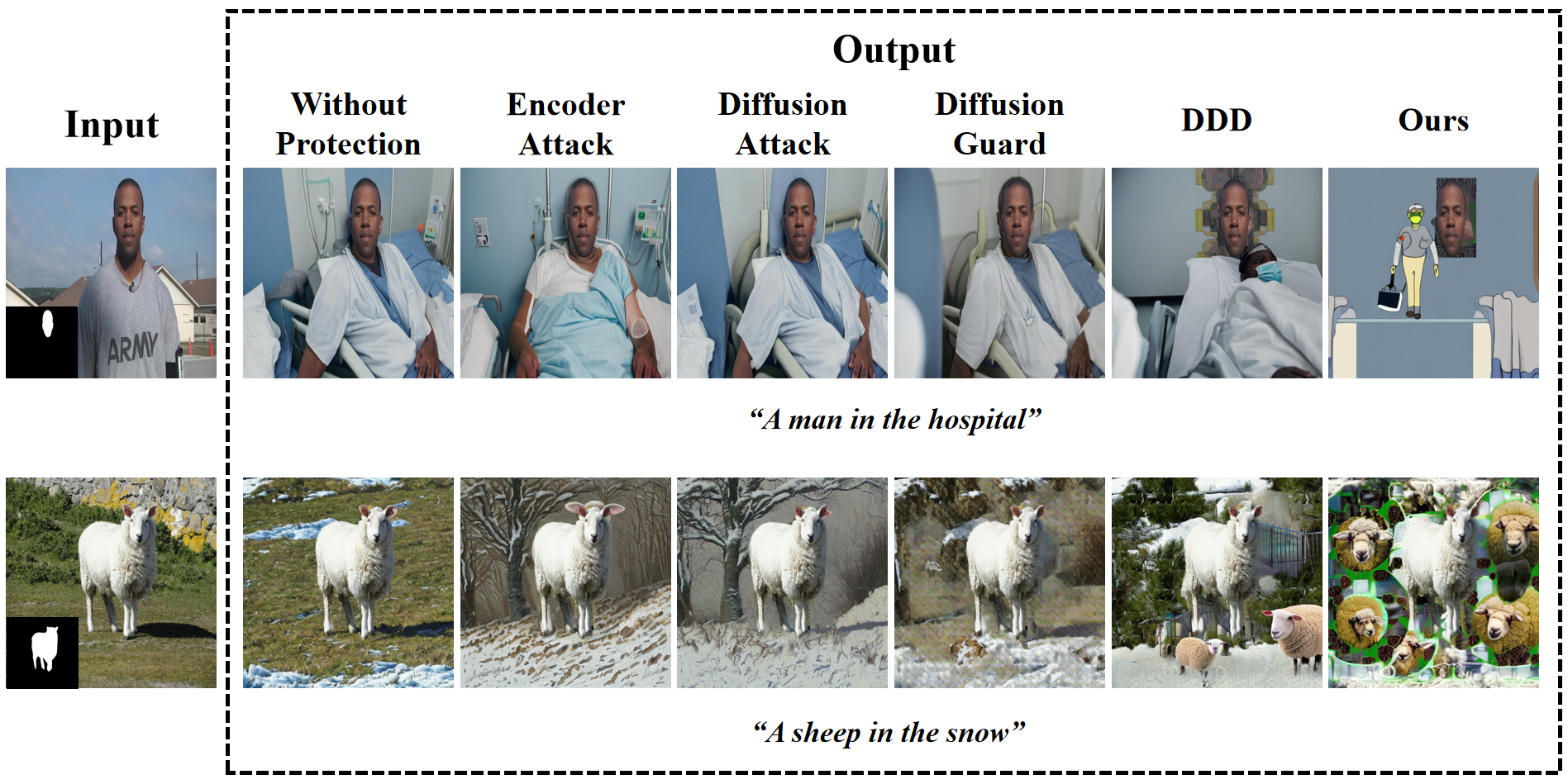}
    \caption{Comparison of our method with baseline approaches. Face protection case with prompt "A man in the hospital" (Top); Instance protection case with prompt "A sheep in the snow" (Bottom).}
    \label{fig:comparison}
\end{figure}
	
Current research on protection targeting Stable Diffusion inpainting tasks remains limited. We experimentally compared the effectiveness of existing open source attack methods, and we demonstrate that our approach achieves state-of-the-art performance. As shown in Figure \ref{fig:comparison}, we evaluate our method against four representative baselines: EncoderAttack and DiffusionAttack from Photoguard \cite{Salman2023Raising}, as well as DiffusionGuard \cite{choi2025diffusionguard} and DDD \cite{son2024disrupting}. The results demonstrate that current methods exhibit limited protective capability (Columns 3-6, Row 2) or complete functional failure (Columns 4-5, Row 1). In stark contrast, as illustrated in row 1 column 7, the SDA-protected region and the generated area are completely decoupled into two distinct images - one depicting a realistic human face and the other presenting a cartoon character. In row 2 column 7, the overall image structure appears disordered, demonstrating that SDA effectively prevents the model from capturing information from the protected region. Consequently, the model is compelled to complete the remaining content without proper guidance, resulting in an incoherent composition where the two components fail to form a logically consistent image when merged. 
	
\begin{table}[t]
\centering
\caption{The performance of SDA and competitors.}
\label{tab:comparison}
    \resizebox{\textwidth}{!}{%
        \begin{tabular}{cccccccc}
            \hline
            & VIF$\downarrow$   & SSIM$\downarrow$  & PSNR$\downarrow$ & FID$\uparrow$   & LPIPS$\uparrow$ & CLIP Score$\downarrow$ & PIQE$\uparrow$           \\ \hline
            \multicolumn{8}{c}{face dataset} \\ \hline
            RandomNoise     & 0.2492          & 0.6385          & 17.12          & 155.45          & 0.4471          & 28.79          & 25.80          \\
            EncoderAttack \cite{Salman2023Raising}   & 0.2196          & 0.5736          & 14.90          & 183.85          & 0.4941          & 29.01          & 30.27          \\
            DiffusionAttack \cite{Salman2023Raising} & 0.2234          & 0.5751          & 14.60          & 189.16          & 0.5293          & 28.01          & 32.36          \\
            DiffusionGuard \cite{choi2025diffusionguard}  & 0.2445          & 0.5895          & 14.88          & 197.89          & 0.5102          & 27.74          & 33.47          \\
            DDD \cite{son2024disrupting}             & 0.1750          & 0.5107          & 13.41          & 233.19          & 0.5797          & 26.01          & 38.92          \\
            Ours            & \textbf{0.1583} & \textbf{0.4733} & \textbf{12.37} & \textbf{265.21} & \textbf{0.6240} & \textbf{25.06} & \textbf{43.02} \\ \hline
            \multicolumn{8}{c}{instance dataset}  \\ \hline
            RandomNoise     & 0.3763          & 0.7215          & 20.32          & 64.62           & 0.2328          & 30.46          & 26.89          \\
            EncoderAttack \cite{Salman2023Raising}   & 0.2546          & 0.5531          & 16.04          & 94.21           & 0.4108          & 30.10          & 30.48          \\
            DiffusionAttack \cite{Salman2023Raising} & 0.2682          & 0.5639          & 15.30          & 106.57          & 0.4248          & 30.17          & 26.66          \\
            DiffusionGuard \cite{choi2025diffusionguard}  & 0.2620          & 0.5744          & 15.64          & 98.52           & 0.4170          & 29.92          & 28.99          \\
            DDD \cite{son2024disrupting}             & 0.1773          & 0.4585          & 12.73          & 150.131         & 0.5214          & 28.04          & 33.05          \\
            Ours            & \textbf{0.1586} & \textbf{0.4242} & \textbf{12.11} & \textbf{179.21} & \textbf{0.5705} & \textbf{27.51} & \textbf{36.97} \\ \hline
        \end{tabular}%
    }
		
\end{table}
	
Table \ref{tab:comparison} presents a quantitative comparison of protection methods across two datasets by IQA PyTorch \cite{pyiqa}, demonstrating our SDA's superior performance. Existing methods show inconsistent results: DiffusionGuard (Row 6, Column 2) achieves minimal VIF improvement in face data (0.2445 vs control 0.2492 at Row 3, Column 1, difference 0.0047), while SDA (Row 8, Column 2) shows significantly better results (0.1583, difference 0.0909). Similarly in instance data, EncoderAttack (Row 11, Column 5) yields FID 94.21 with limited difference from the control (64.62 at Row 10, Column 5, difference 29.59), whereas SDA (Row 15, Column 5) delivers substantially stronger protection (179.21, difference 114.59).
	
% \subsection{Transferability}
	
\subsection{Robustness analysis}
	
To validate real-world applicability, we systematically evaluate SDA's robustness across multiple dimensions, including data augmentation strategies, different model versions. We also assess the performance of various protection methods under mask augmentation \cite{choi2025diffusionguard} conditions that simulate real-world deployment scenarios. Due to the page limitation, we evaluate our SDA under different hyperparameter configurations in Appendix. 
	
\begin{figure}[t]
    \centering
    \includegraphics[width=0.7\textwidth]{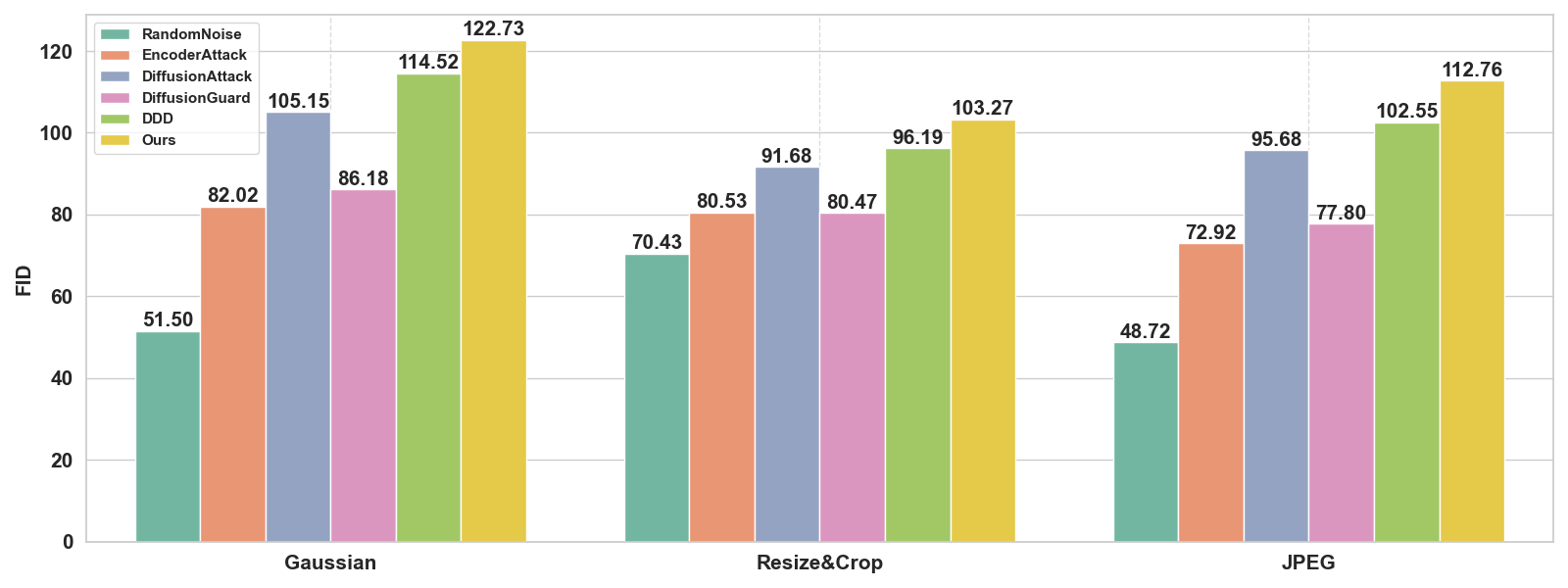}
    \caption{Performance comparison of protection methods under data augmentation. We compute FID on the instance dataset across different protection methods.}
    \label{fig:robustness}
\end{figure}

\paragraph{Data augmentations} As demonstrated in Figure \ref{fig:robustness}, our proposed SDA exhibits notable robustness against common data augmentation operations, including Gaussian noise addition, random post-resize cropping, and JPEG compression, while maintaining its state-of-the-art protection performance. For example, SDA achieves an FID of 122.73 (yellow bar in the left subfigure), demonstrating a 71.23 increase over the control group (51.50, green bar), a significantly larger gap than EncoderAttack's marginal 30.52 difference (82.02, orange bar).
	
\paragraph{Transferability} Transferability measures the effectiveness of protective perturbations optimized for one model when applied to other models. We evaluate this property using two distinct inpainting checkpoints: RunwayML's v1.5 and StabilityAI's v2.0. In particular, while both checkpoints share the same network architecture, they represent independent implementations with different training protocols. As specified in our methodology, since the protective perturbations were optimized on v2.0, attacks against this version constitute white-box scenarios, whereas attacks against v1.5 represent black-box conditions.
	
\begin{table}[t]
    \centering
    \caption{Quantitative evaluation of SDA's black-box transferability.}
    \label{tab:transferability}
    \resizebox{0.9\textwidth}{!}{%
        \begin{tabular}{cccccccc}
            \hline
            & VIF$\downarrow$   & SSIM$\downarrow$  & PSNR$\downarrow$ & FID$\uparrow$   & LPIPS$\uparrow$ & CLIP Score$\downarrow$ & PIQE$\uparrow$  \\ \hline
            \multicolumn{8}{c}{v2}                                                       \\ \hline
            RandomNoise & 0.2492 & 0.6385 & 17.12 & 155.45 & 0.4471 & 28.79      & 25.80 \\
            Ours        & \textbf{0.1583} & \textbf{0.4733} & \textbf{12.37} & \textbf{265.21} & \textbf{0.6240} & \textbf{25.06}      & \textbf{43.02} \\ \hline
            \multicolumn{8}{c}{v1.5}                                                     \\ \hline
            RandomNoise & 0.2684 & 0.6591 & 17.49 & 141.56 & 0.3994 & 28.64      & 23.35 \\
            Ours        & \textbf{0.1726} & \textbf{0.5022} & \textbf{13.30} & \textbf{222.21} & \textbf{0.5684} & \textbf{27.78}      & \textbf{39.46} \\ \hline
        \end{tabular}%
    }
\end{table}
	
We evaluate transferability on the face dataset, with Table \ref{tab:transferability} quantifying SDA's black-box protection performance. Our method demonstrates consistent effectiveness against unknown threat models, as evidenced by PIQE scores of 39.46 for SDA-protected images versus 23.35 for the control group (Row 7, Column 8) when tested on the Stable Diffusion v1.5 inpainting model.
	
\begin{figure}[t]
    \centering
    \includegraphics[width=0.8\textwidth]{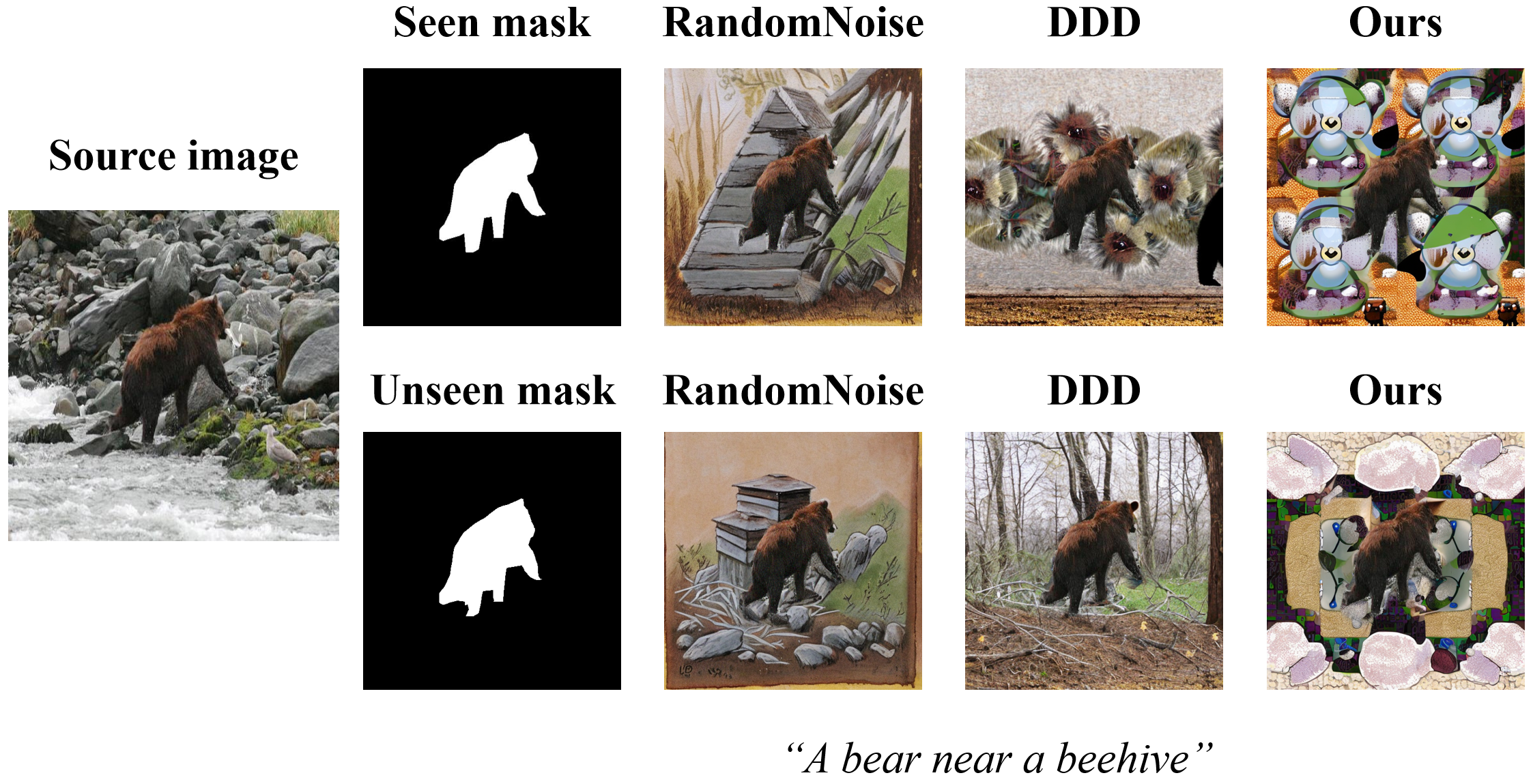}
    \caption{The visualization performance of our SDA under seen and unseen mask conditions. Seen masks (used for perturbation optimization) versus unseen masks (augmented variants). }
    \label{fig:robust_mask}
\end{figure}

\paragraph{Mask augmentation} To ensure the applicability in the real world, we evaluate the inevitable discrepancy between the masks specified by the attacker and those used during perturbation optimization. Following Choi et al.'s methodology \cite{choi2025diffusionguard}, we implement mask augmentation to simulate various "unseen" masking scenarios likely encountered in practice, as visualized in Figure \ref{fig:robust_mask}, which exhibits more challenging characteristics, including coarser boundaries and irregular shapes.
	
Our SDA demonstrates superior cross-mask robustness: while DDD shows effective protection only for seen masks (Row 1, Column 3), it fails completely against unseen masks (Row 2, Column 3). In contrast, SDA maintains consistently high performance for both mask types (Rows 1-2, Column 4), validating its practical applicability.
	
\begin{table}[t]
    \centering
    \caption{The performance of SDA and competitors under unseen mask on the instance dataset.}
    \label{tab:robust_mask}
    \resizebox{\textwidth}{!}{%
        \begin{tabular}{cccccccc}
            \hline
            & VIF$\downarrow$   & SSIM$\downarrow$  & PSNR$\downarrow$ & FID$\uparrow$   & LPIPS$\uparrow$ & CLIP Score$\downarrow$ & PIQE$\uparrow$ \\ \hline
            RandomNoise     & 0.3810 & 0.7248 & 20.50 & 65.71  & 0.2325 & 30.41       & 25.48 \\
            EncoderAttack \cite{Salman2023Raising}   & 0.2533 & 0.5670 & 16.10 & 92.77  & 0.3885 & 30.17       & 29.70 \\
            DiffusionAttack \cite{Salman2023Raising} & 0.2651 & 0.5798 & 15.88 & 94.67  & 0.3975 & 30.07       & 28.32 \\
            DiffusionGuard \cite{choi2025diffusionguard}  & 0.2660 & 0.5790 & 15.87 & 95.65  & 0.4035 & 30.02       & 28.57 \\
            DDD \cite{son2024disrupting}             & 0.2056 & 0.5160 & 13.95 & 117.28 & 0.4574 & 29.64       & 32.32 \\
            Ours & \textbf{0.1968} & \textbf{0.4967} & \textbf{13.48} & \textbf{126.65} & \textbf{0.4838} & \textbf{29.12} & \textbf{33.32} \\ \hline
        \end{tabular}%
    }
\end{table}
	
Table \ref{tab:robust_mask} shows the unseen-mask performance of SDA, our method maintains superior effectiveness in this challenging setting, evidenced by a PIQE score of 33.32 (Row 7, Column 8), representing a 7.84-point degradation from the control group (25.48, Row 2, Column 8) and significantly larger deviation than DiffusionAttack's marginal 2.84-point difference (28.32 vs. 25.48, Row 4, Column 8).

\section{Conclusion}
	
This paper presents Structure Disruption Attack (SDA), a novel protection framework for safeguarding sensitive image regions against diffusion-based inpainting. Building upon the coarse-to-fine generation paradigm of diffusion models, SDA strategically disrupts the self-attention mechanism during initial denoising steps, effectively compromising the model's structural generation capability and preventing coherent output synthesis. Our attention map visualizations provide compelling evidence for the proposed mechanism, clearly demonstrating how SDA's targeted interference disrupts critical attention patterns essential for proper image composition. Extensive experiments demonstrate that SDA achieves state-of-the-art protection performance while exhibiting remarkable robustness across: (1) various image augmentations, (2) different model versions, (3) diverse hyperparameter configurations, and (4) varying mask sizes and text prompts. These advantages establish SDA as a reliable defensive solution against potential misuse of diffusion models.
	
\newpage
	
%%%%%%%%%%%%%%%%%%%%%%%%%%%%%%%%%%%%%%%%%
\bibliographystyle{unsrtnat}
\bibliography{main}
%%%%%%%%%%%%%%%%%%%%%%%%%%%%%%%%%%%%
% \section*{References}

% References follow the acknowledgments in the camera-ready paper. Use unnumbered first-level heading for
% the references. Any choice of citation style is acceptable as long as you are
% consistent. It is permissible to reduce the font size to \verb+small+ (9 point)
% when listing the references.
% Note that the Reference section does not count towards the page limit.
% \medskip

% {
	% \small

	% [1] Alexander, J.A.\ \& Mozer, M.C.\ (1995) Template-based algorithms for
	% connectionist rule extraction. In G.\ Tesauro, D.S.\ Touretzky and T.K.\ Leen
	% (eds.), {\it Advances in Neural Information Processing Systems 7},
	% pp.\ 609--616. Cambridge, MA: MIT Press.

	% [2] Bower, J.M.\ \& Beeman, D.\ (1995) {\it The Book of GENESIS: Exploring
			%   Realistic Neural Models with the GEneral NEural SImulation System.}  New York:
	% TELOS/Springer--Verlag.

	% [3] Hasselmo, M.E., Schnell, E.\ \& Barkai, E.\ (1995) Dynamics of learning and
	% recall at excitatory recurrent synapses and cholinergic modulation in rat
	% hippocampal region CA3. {\it Journal of Neuroscience} {\bf 15}(7):5249-5262.
	% }

%%%%%%%%%%%%%%%%%%%%%%%%%%%%%%%%%%%%%%%%%%%%%%%%%%%%%%%%%%%%
\newpage
\appendix
	
\section{Appendix}
\subsection{Additional experiments}
\label{Additional Experiments}
\paragraph{Computational efficiency} 

The design rationale of SDA demonstrates its computational superiority over existing methods. Our approach simplifies the attack process by targeting only the initial denoising steps, as opposed to the full-chain attack required by Diffusion Attack. Compared with DDD which necessitates prior optimization of prompt embeddings for optimal matching, SDA directly employs null-text prompts, effectively eliminating the need for prompt embedding optimization. This strategic simplification not only enhances protection efficiency, but also maintains comparable protection performance.

\begin{figure}[h]
    \centering
    \includegraphics[width=\textwidth]{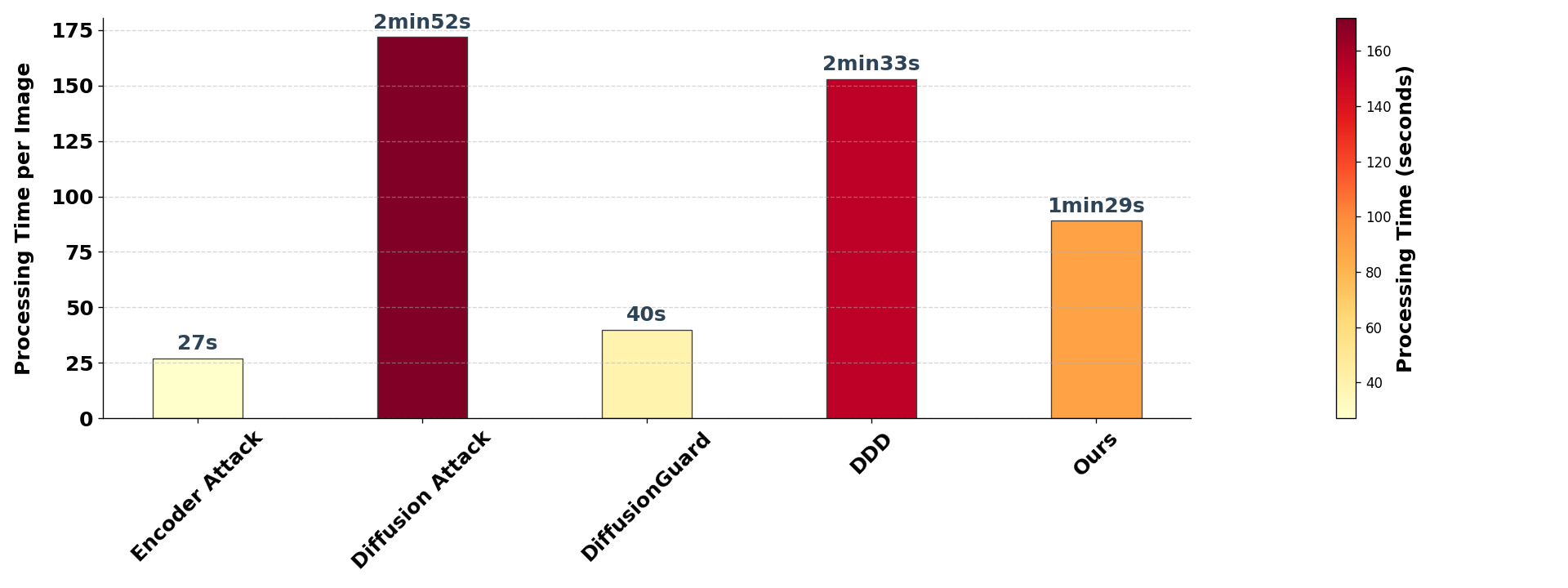}
    \caption{Time cost of different protection methods.}
    \label{fig:time}
\end{figure}

To ensure experimental fairness, we conducted time measurements under strictly controlled conditions: gradient repetition steps (grad\_reps) were uniformly set to 1 and iteration counts (iters) fixed at 300 for all methods. As shown in Figure \ref{fig:time}, the recorded protection durations reveal significant efficiency improvements: Diffusion Attack required 2 minutes 52 seconds per image, DDD consumed 2 minutes 33 seconds, while our SDA achieved the task in merely 1 minute 29 seconds – nearly twice as fast as full-chain approaches.

Notably, our comparative analysis intentionally excluded performance considerations to focus purely on computational efficiency. When optimizing for maximum protection performance (e.g., using DiffusionGuard's official recommendation of 800 iterations), the processing time increases to 1 minute 46 seconds, still surpassing SDA's 1 minute 29 seconds execution at 300 iterations. Crucially, even with reduced iterations, SDA maintains superior protection efficacy as demonstrated in our security evaluation experiments.

\paragraph{The sensitivity to the hyperparameters of inpainting} In Stable Diffusion inpainting tasks, hyperparameter selection critically influences the generation outcomes, particularly the $strength$ parameter which governs the dependency on source images. The parameter exhibits a continuous spectrum of control: at $strength=1$, the generation becomes completely stochastic, while $strength=0$ forces strict adherence to the original image content. We analyze how protective efficacy varies across this parameter continuum.

\begin{figure}[h]
    \centering
    \includegraphics[width=\textwidth]{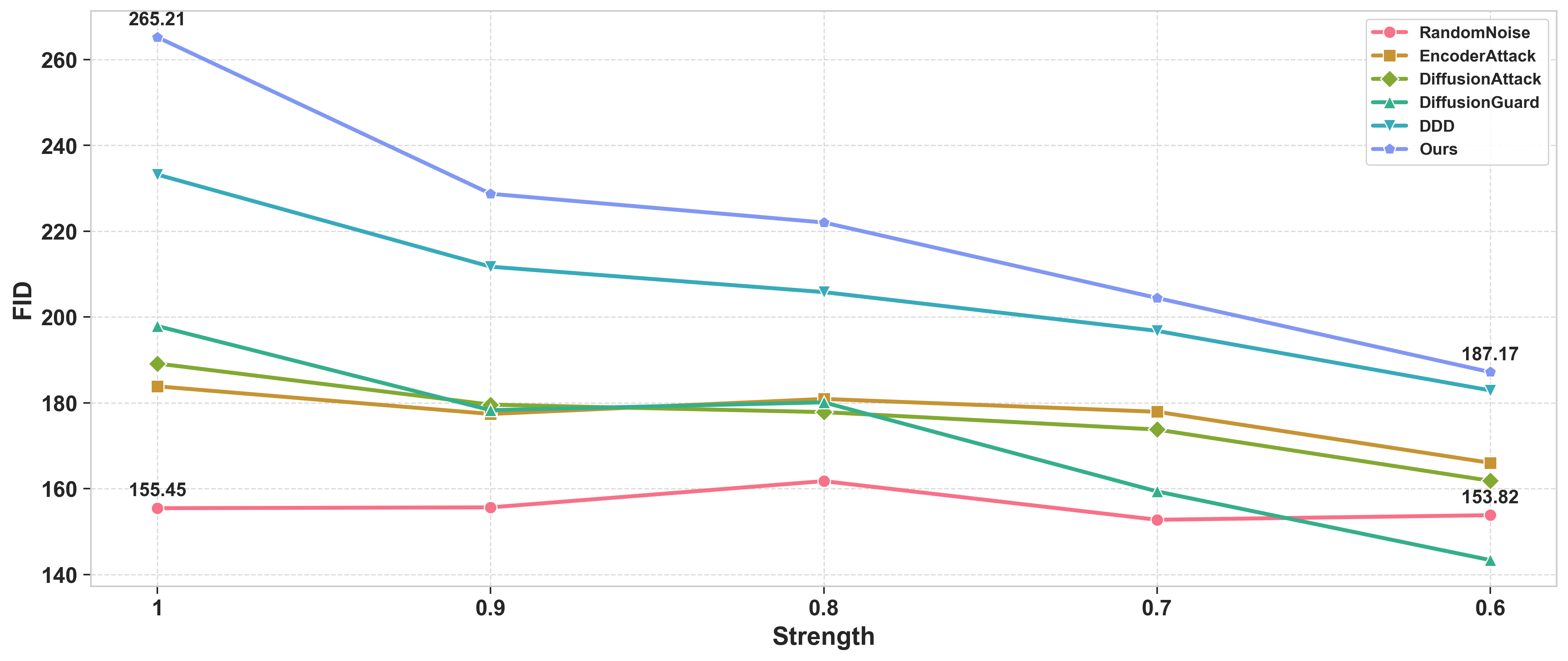}
    \caption{Performance variation of protection methods with inpainting strength. Evaluation conducted on the face dataset by systematically varying the inpainting strength parameter, with FID scores recorded for each configuration. The RandomNoise condition (consistent with prior experimental settings) serves as the control baseline.}
    \label{fig:strength}
\end{figure}
	
Fig. \ref{fig:strength} demonstrates that protective efficacy generally diminishes with decreasing inpainting strength, as the original image content progressively dominates the generation process, thereby attenuating adversarial perturbations. Notably, our method maintains significant effectiveness even at $strength=0.6$ (purple curve), while competing approaches like DiffusionGuard (cyan curve) show near-complete performance degradation by $strength=0.7$. Most baseline methods (yellow/green/cyan curves) already fail at $strength=0.6$, collectively indicating that SDA exhibits superior robustness to inpainting hyperparameter variations compared to existing solutions.

\subsection{Broader impacts and limitations}
\label{Limitation}
\paragraph{Broader impacts} Existing approaches to prevent the misuse of diffusion models primarily focus on maximizing denoising loss or disrupting cross-attention modules that govern text-image alignment. In contrast, our method innovatively targets the self-attention mechanism. Through empirical analysis, we demonstrate that this attack not only prevents the model from capturing structural contour information but also triggers a cascading effect, leading to the subsequent failure of text-alignment capabilities. From a technical perspective, our findings highlight the critical role of self-attention in diffusion-based generative models, calling for increased community attention to its vulnerabilities and robustness. In practical terms, the proposed method provides a novel defense mechanism to safeguard user images against unauthorized malicious edits, thereby contributing to the development of safer and more ethical AI applications.

\paragraph{Limitation} While our study provides novel insights into defending against diffusion inpainting-based image editing, its scope is currently limited to this specific attack scenario. Notably, modern image editing increasingly relies on instruction-driven methods (e.g., DiffEdit \cite{couairon2023diffedit} for text-guided manipulation and MasaCtrl \cite{cao_2023_masactrl} for fine-grained control over latent space), where malicious edits can be implemented without explicit inpainting masks. Our framework has not yet been systematically evaluated in these emerging scenarios, potentially restricting its generalizability to broader attack surfaces. Furthermore, the rapid evolution of image editing techniques (e.g., zero-shot editing and prompt engineering) necessitates continuous adaptation of defense strategies. Addressing these gaps will be a critical focus of our future research.

%%%%%%%%%%%%%%%%%%%%%%%%%%%%%%%%%%%%%%%%%%%%%%%%%%%%%%%%%%%%

\end{document}